\documentclass[sn-mathphys,Numbered]{sn-jnl}

\usepackage{graphicx}%
\usepackage{multirow}%
\usepackage{amsmath,amssymb,amsfonts}%
\usepackage{amsthm}%
\usepackage{mathrsfs}%
\usepackage[title]{appendix}%
\usepackage{xcolor}%
\usepackage{textcomp}%
\usepackage{manyfoot}%
\usepackage{booktabs}%
\usepackage{algorithm}%
\usepackage{algorithmicx}%
\usepackage{algpseudocode}%
\usepackage{listings}%

\usepackage{multirow}
\usepackage{array}
\usepackage{vcell}
\usepackage{balance}
\newcolumntype{C}[1]{>{\centering\arraybackslash}p{#1}}
\usepackage{enumitem}
\UseRawInputEncoding

\theoremstyle{thmstyleone}%
\theoremstyle{thmstyletwo}%

\theoremstyle{thmstylethree}%

\begin{document}

\title[Article Title]{An Experimental Study on Data Augmentation Techniques for Named Entity Recognition on Low-Resource Domains}


\author*[1]{\fnm{Arthur} \sur{Elwing Torres}}\email{arthur.torres@icomp.ufam.edu.br}

\author[1]{\fnm{Edleno} \sur{Silva de Moura}}\email{edleno@icomp.ufam.edu.br}

\author[1]{\fnm{Altigran} \sur{Soares da Silva}}\email{alti@icomp.ufam.edu.br}

\author[2]{\fnm{Mario} \sur{A. Nascimento}}\email{m.nascimento@northeastern.edu}

\author[3]{\fnm{Filipe} \sur{Mesquita}}\email{filipe@diffbot.com}

\affil*[1]{\orgname{Universidade Federal do Amazonas}, \city{Manaus}, \state{Amazonas}, \country{Brazil}}

\affil[2]{\orgname{Northeastern University}, \city{Vancouver}, \state{BC}, \country{Canada}}

\affil[3]{\orgname{Diffbot Technologies Corp.}, \city{Menlo Park}, \state{California}, \country{USA}}


\abstract{Named Entity Recognition (NER) is a machine learning task that traditionally relies on supervised learning and annotated data. Acquiring such data is often a challenge, particularly in specialized fields like medical, legal, and financial sectors. Those are commonly referred to as \emph{low-resource} domains, which comprise \emph{long-tail} entities, due to the scarcity of available data. To address this, data augmentation techniques are increasingly being employed to generate additional training instances from the original dataset. In this study, we evaluate the effectiveness of two prominent text augmentation techniques, \emph{Mention Replacement} and \emph{Contextual Word Replacement}, on two widely-used NER models, Bi-LSTM+CRF and BERT. We conduct experiments on four datasets from low-resource domains, and we explore the impact of various combinations of training subset sizes and number of augmented examples. We not only confirm that data augmentation is particularly beneficial for smaller datasets, but we also demonstrate that there is no universally optimal number of augmented examples, i.e., NER practitioners must experiment with different quantities in order to fine-tune their projects. 
}




\maketitle

\section{Introduction}\label{introduction}

Named Entity Recognition (NER) is a specialized subtask of Information Extraction that focuses on identifying and categorizing mentions of real-world entities in text~\cite{nlp-book-jurafsky}. Common entity types such as \emph{Person}, \emph{Location}, and \emph{Organization} are frequently encountered in abundant online sources such as news, websites, and social media, and thus form what are called \emph{generic} domains. These domains have been extensively studied, and numerous datasets are readily available for research (e.g.,~\cite{automated-concatenation,li-etal-2020-dice}). In contrast, entities in specialized application domains, also called \emph{low-resource} domains~\cite{dai-adel-2020-analysis,ijcai2022p590}, like medical, legal, and financial sectors, are termed as \emph{long-tail} entities~\cite{10.1145/3366423.3380123,Liu2020LongtailDE}. These domains are less commonly found in typical online text corpora. Additionally, the annotation of these specialized entities often demands technical expertise, making the generation of annotated datasets a challenging task~\cite{hou-etal-2020-shot,friedrich-etal-2020-sofc}. Given the critical role of annotated data in the development, training, and testing of NER models~\cite{devlin-etal-2019-bert,DBLP:journals/corr/HuangXY15}, there is a growing interest in methodologies for building NER models in scenarios where annotated data are scarce or difficult to generate, as is the case with these low-resource domains~\cite{book-machine-learning}.

One such methodology is the application of data augmentation, which is a technique that improves the construction of effective NER models when annotated training data is limited or costly to obtain. This approach automatically generates new data samples by applying transformations to existing data~\cite{article-da-warp, article-survey-da, article-imagenet}. Originally popularized in the field of computer vision, data augmentation is gaining traction in Natural Language Processing (NLP) tasks, including NER~\cite{dai-adel-2020-analysis,Liu2020LongtailDE,ijcai2022p590,10.1007/978-3-030-51310-8_2,Neural-Cross-Lingual}. For example, a sentence like \emph{I have a cat named Serena} could be transformed into \emph{You have a dog named Beethoven} or \emph{I own a cat called Serena}, thereby enriching the training set. These augmented sentences maintain the core structure of the original, while introducing variations that enhance the diversity of the training set.

Despite the growing interest, existing research has not sufficiently explored how the volume of augmented examples influences the performance of NER models. This is a crucial parameter that needs to be determined prior to model training. In this study, we examine the effectiveness of two specific data augmentation techniques, \emph{Mention Replacement} (MR) and \emph{Contextual Word Replacement} (CWR), across four datasets from various low-resource domains. 

We experiment with different quantities of augmented sentences and diverse samples from these datasets, employing BERT~\cite{devlin-etal-2019-bert} and Bi-LSTM+CRF~\cite{DBLP:journals/corr/HuangXY15}, two widely-used NER architectures~\cite{dai-adel-2020-analysis, ijcai2022p590}.
Consistent with prior research, our findings confirm that data augmentation is particularly beneficial for smaller datasets. However, for larger datasets, we observed that models trained with augmented data yielded equivalent or even inferior performance compared to those trained without augmentation. We examine the impact of varying amounts of augmented data on model performance as our key contribution. We demonstrate that there exists a saturation point beyond which additional augmentation may degrade model quality. This underscores the need for NER practitioners to carefully experiment with different volumes of augmented examples, as the optimal quantity cannot be predetermined.

We further contribute to the field by evaluating NER models on a previously unstudied dataset, which will be publicly released along with the code used in this study. Our experiments also reveal that \emph{Contextual Word Replacement} (CWR) generally outperforms \emph{Mention Replacement} (MR), and that BERT models benefit more from data augmentation than Bi-LSTM+CRF models, at least within the scope of our datasets.

The remainder of this paper is structured as follows: Section~\ref{sec:background} provides an overview of NER models and text data augmentation techniques. Section~\ref{sec:related-work} reviews prior works that have applied data augmentation to NER. Section~\ref{sec:exp-setup} details the datasets, augmentation techniques, and experimental parameters. Finally, Section~\ref{sec:results} presents and discusses our findings, and Section~\ref{sec:conclusion} concludes the paper.

\section{Background}\label{sec:background}

In this section, we present a review of NER techniques in the literature, as well as a summary of best-known text augmentation methods, focusing mainly on the ones applied alongside those techniques.

\subsection{NER Techniques}

In the realm of Named Entity Recognition (NER), various architectures have been proposed in the literature~\cite{9039685, yadav-bethard-2018-survey}. Classical approaches include the \emph{Hidden Markov Model} (HMM) \cite{10.1214/aoms/1177699147}, the \emph{Maximum-entropy Markov Model} (MEMM) \cite{10.5555/645529.658277}, and the \emph{Conditional Random Fields} (CRF) \cite{10.5555/645530.655813}. In addition to these, Deep Learning-based models such as \emph{Long short-term Memory} (LSTM) \cite{10.1162/neco.1997.9.8.1735} and \emph{Gated Recurrent Unit} (GRU)~\cite{cho-etal-2014-learning} have also emerged as effective architectures for NER tasks. This diverse range of models offers a rich set of techniques for developing robust and accurate NER systems.

Two prominent architectures have shown exceptional performance in within the context of NER, particularly in low-resource settings~\cite{dai-adel-2020-analysis, ijcai2022p590}. The first one, \emph{Bi-LSTM+CRF}~\cite{DBLP:journals/corr/HuangXY15}, is a deep neural network that incorporates an embedding layer, followed by dual bidirectional \emph{LSTM} layers to capture both forward and backward semantic dependencies, and concludes with a \emph{CRF} layer for sequence tagging. The second one, \emph{BERT} (Bidirectional Encoder Representations from Transformers)~\cite{devlin-etal-2019-bert}, is an evolution of the \emph{Transformer} architecture, utilizing an attention mechanism to process an entire sequence of words simultaneously and understand their contextual relationships. Comparative studies often reveal that BERT models outperform Bi-LSTM+CRF models in NER tasks~\cite{9039685}. However, it is important to consider the higher computational overhead associated with BERT, typically involving the estimation of approximately 110 million parameters, compared to a typically lower parameter count for Bi-LSTM+CRF models.

\subsection{Text Augmentation Methods}

We summarize some of the best-known text augmentation techniques in the literature below. Table~\ref{tab:aug-techs} shows examples of applying these techniques. We adopt the well-known IOB~\cite{ramshaw-marcus-1995-text} tagging scheme.  In this scheme, each target entity is tagged with its corresponding entity type, prefixed by "B-". Tokens that do not belong to any relevant entity type are labeled as "O" and remain unannotated. For sequences of consecutive tokens that share the same entity type, the initial token is prefixed with "B-", while subsequent tokens receive an "I-" prefix to denote a contiguous "chunk" of tokens.
\newline
\begin{description}
\setlength\itemsep{0.5em}
    \item[Easy Data Augmentation (EDA)~\cite{easy-da}] EDA encompasses a suite of straightforward text manipulations, which include: 
    \begin{itemize}
        \item Synonym Replacement (SR) - Replacing random words in a sentence with their corresponding synonyms from a thesaurus.
        \item Random Insertion (RI) - Inserting random synonyms of words at arbitrary positions in the sentence.
        \item Random Swap (RS) - Interchanging the positions of two randomly-selected words in the sentence.
        \item Random Deletion (RD) - Removing a randomly-chosen word from the sentence.
    \end{itemize}
    Notably, some techniques discussed later in this section can be viewed as nuanced variations of EDA methods.
    
    \item[Label-wise Token Replacement (LwTR)~\cite{dai-adel-2020-analysis}] This method involves replacing random words in a sentence with other words from the original training set that share the same label.
    
    \item[Mention Replacement (MR)~\cite{dai-adel-2020-analysis}] An extension of LwTR, MR replaces randomly-selected mentions, i.e., one or more contiguous tokens with a uniform label type, with other mentions from the original training set that have the same label type. The number of tokens in the replacement mention may differ from the original.

    \item[Shuffle Within Segments (SWS)~\cite{dai-adel-2020-analysis}] In this approach, a sentence is segmented based on mentions (as defined in MR). Tokens within each segment are then randomly reordered.

    \item[Entity Replacement (ER)~\cite{Liu2020LongtailDE}] A variant of MR, ER replaces entities with alternative entities from sources other than the original training set.

    \item[Contextual Word Replacement (CWR)~\cite{jiao-etal-2020-tinybert}] This method employs BERT models to replace words contextually within a sentence.

    \item[Entity Mask (EM)~\cite{Liu2020LongtailDE}] Similar to CWR but focuses exclusively on replacing entity tokens.

    \item[Label-conditioned Word Replacement (LcWR)~\cite{ijcai2022p590}] This technique utilizes a pre-trained BERT model that is fine-tuned to capture word-label dependencies for word replacement tasks.

    \item[Inner and Outer Data Augmentation~\cite{10.1007/978-3-030-51310-8_2}] Involves the replacement of non-specific entity mentions, termed as \emph{descriptors}, with proper names to enhance specificity (e.g., replacing "medicine" with "Paracetamol"). The \emph{Inner} variant applies this replacement to sentences in the original training set, whereas the \emph{Outer} variant extends this to automatically annotate unlabeled sentences that contain descriptors and are outside the original dataset.
\end{description}

\begin{sidewaystable}
    \caption{Text Augmentation Techniques with examples}
    \footnotesize\setlength{\tabcolsep}{4pt} 
    \label{tab:aug-techs}
        \begin{tabular}{p{0.5cm}p{0.5cm}p{0.7cm}p{2cm}p{2cm}p{0.6cm}p{1.45cm}p{1.45cm}p{1.45cm}p{1.35cm}p{0.1cm}p{3cm}}
            \toprule
            \multicolumn{1}{c}{\multirow{2}{*}{\textbf{Technique}}} & \multicolumn{10}{c}{\multirow{2}{*}{\textbf{Sentence}}} & \multicolumn{1}{c}{\multirow{2}{*}{\textbf{Obs.}}} \\
            \multicolumn{1}{c}{} & \multicolumn{10}{c}{} & \multicolumn{1}{c}{} \\
            \midrule
            \multirow{2}{*}{None} & I & took & a & medicine & today & for & acute & colitis & . &  &  \\
             & O & O & O & B-treatment & O & O & B-disease & I-disease & O &  &  \\
            \midrule
            \multirow{2}{*}{EDA - SR} & I & took & a & \textbf{remedy} & today & for & acute & colitis & . &  &  \\
             & O & O & O & B-treatment & O & O & B-disease & I-disease & O &  &  \\
            \midrule
            \multirow{2}{*}{EDA - RI} & I & took & a & medicine & today & \textbf{and} & for & acute & colitis & . &  \\
             & O & O & O & B-treatment & O & O & O & B-disease & I-disease & O &  \\
            \midrule
            \multirow{2}{*}{EDA - RS} & I & \textbf{today} & a & medicine & \textbf{took} & for & acute & colitis & . &  &  \\
             & O & O & O & B-treatment & O & O & B-disease & I-disease & O &  &  \\
            \midrule
            \multirow{2}{*}{EDA - RD} & I & took & medicine & today & for & acute & colitis & . &  &  & \\
             & O & O & B-treatment & O & O & B-disease & I-disease & O &  &  & Removed token "a" \\
            \midrule
            \multirow{2}{*}{LwTR} & I & \textbf{he} & a & \textbf{surgery} & today & for & acute & \textbf{headache} & . &  &  \\
             & O & O & O & B-treatment & O & O & B-disease & I-disease & O &  & New tokens came from training corpus \\
            \midrule
            \multirow{2}{*}{MR} & I & took & a & medicine & today & for & \textbf{cancer} & . &  &  & \\
             & O & O & O & B-treatment & O & O & B-disease & O &  &  & New tokens came from training corpus \\
            \midrule
            \multirow{2}{*}{SWS} & \textbf{took} & \textbf{a} & \textbf{I} & medicine & \textbf{for} & \textbf{today} & \textbf{colitis} & \textbf{acute} & . &  &  \\
             & O & O & O & B-treatment & O & O & B-disease & I-disease & O &  &  \\
            \midrule
            \multirow{2}{*}{ER} & I & took & a & medicine & today & for & \textbf{Alzheimer} & \textbf{syndrome} & . &  & \\
             & O & O & O & B-treatment & O & O & B-disease & I-disease & O &  & New tokens came from external sources \\
            \midrule
            \multirow{2}{*}{EM} & I & took & a & \textbf{car} & today & for & \textbf{country} & . &  &  & \\
             & O & O & O & B-treatment & O & O & B-disease & O &  &  & New tokens are generated randomly \\
            \midrule
            \multirow{2}{*}{CWR} & I & \textbf{got} & a & medicine & \textbf{now} & for & acute & colitis & . &  &  \\
             & O & O & O & B-treatment & O & O & B-disease & I-disease & O &  & New tokens came from BERT \\
            \midrule
            \multirow{2}{*}{LcWR} & I & took & a & \textbf{surgery} & today & for & acute & \textbf{headache} & . &  & \\
             & O & O & O & B-treatment & O & O & B-disease & I-disease & O &  & New tokens came from BERT \\
            \midrule
            \multirow{2}{*}{Inner/Outer} & I & took & a & \textbf{Paracetamol} & today & for & acute & colitis & . &  & \\
             & O & O & O & B-treatment & O & O & B-disease & I-disease & O &  & New tokens came from external sources \\
            \bottomrule
        \end{tabular}
\end{sidewaystable}

In the field of Named Entity Recognition (NER), Bi-LSTM+CRF and BERT have been identified as particularly effective architectures for low-resource domains~\cite{dai-adel-2020-analysis, ijcai2022p590}. To both adhere to established research and examine the cost-effectiveness of data augmentation across different model types, we opted to utilize both architectures. For data augmentation, we focused on methods that are well regarded in the literature and have demonstrated improvements in the performance of the NER model~\cite{dai-adel-2020-analysis,Liu2020LongtailDE,jiao-etal-2020-tinybert,ijcai2022p590}, while also being relatively simple to implement. Hence, we chose Mention Replacement (MR) for augmenting tokens tagged with specific entity types, excluding those labeled "O," and Contextual Word Replacement (CWR) for tokens specifically tagged as "O." Detailed information on the implementation of these augmentation techniques is available in Section~\ref{sec:exp-setup}.

\section{Related Work}\label{sec:related-work}

Next, we discuss previous work on data augmentation for NER models in this section.
One particular issue with current data augmentation methods in the literature is that most of the work done on NLP tasks considers annotations at the sentence level rather than at the token level, as it should be the case with NER~\cite{dai-adel-2020-analysis}. 

In a closely related study, Dai and Adel~\cite{dai-adel-2020-analysis} investigated the effectiveness of four text augmentation techniques, LwTR, SR, MR, and SWS (Section~\ref{sec:background}), on two specific low-resource domains: biomedical and materials science. These authors assessed the impact of text augmentation on NER models employing both Bi-LSTM+CRF and BERT architectures. Their findings indicate that none of the augmentation methods consistently outperforms the others, and that augmentation is more effective when applied to smaller (and random) subsets rather than to the entire dataset. Another relevant contribution comes from Liu et al.~\cite{ijcai2022p590}, who introduced a novel text augmentation technique, LcWR, as well as an automatic sentence annotation approach using BERT, termed \emph{prompting with question answering}. Additionally, they proposed a noise-reduction system to counteract the potential introduction of noisy data by the augmentation techniques. Their experiments, conducted on three datasets, including one in the low-resource domain of materials science, utilized both Bi-LSTM+CRF and BERT architectures. They found that their techniques not only improved the overall model quality but were particularly effective on smaller subsets of the datasets, while also generating more label-consistent augmented data.

Our work aligns with these studies in terms of the experimental protocol, as we too explore various text augmentation techniques across multiple low-resource domains, employing the same NER architectures. However, a key distinction lies in our exploration of different quantities of augmented instances, enabling us to evaluate the impact of these varying amounts on the quality of the NER models.

 In the work of Liu et al.~\cite{Liu2020LongtailDE}, a comprehensive examination of text augmentation techniques, namely ER and EM (Section~\ref{sec:background}), was conducted on a dataset encompassing long-tail entities within academic papers. The authors evaluated the impact of varying data augmentation scales: 20\%, 50\%, and 100\% of the original dataset. This evaluation was performed using a model that seamlessly integrated BERT, Bi-LSTM, and CRF architectures. Their findings demonstrated an enhancement in model quality attributed to both ER and EM augmentation techniques. However, a noteworthy observation is the non-linear relationship between the volume of augmented examples and model improvement. This insight underscores the necessity for further exploration in this domain, as increasing the quantity of augmented data does not invariably bolster the model's performance. Remarkably, this study stands as a pioneering contribution to the literature, being the sole research to rigorously assess the influence of diverse data augmentation magnitudes on NER models to our knowledge.

In their work, Tikhomirov et al.~\cite{10.1007/978-3-030-51310-8_2} introduce two innovative text augmentation techniques, termed "inner" and "outer" descriptor replacement. These methods are meticulously applied to a dataset within the cybersecurity domain, a field characterized by its low-resource nature. However, the application is confined to a limited set of entity types, specifically \emph{virus} and \emph{hacker}. The authors carried out on a comprehensive evaluation, assessing the impact of these augmentation techniques on models constructed using diverse architectures, including CRF, BERT, and two specially fine-tuned variations of BERT, denominated as \emph{RuBERT} and \emph{RuCYBERT}. These tailored models are explicitly designed to address the problem at hand. The empirical results indicate that while the introduced augmentation techniques enhance the model recall, a concomitant decline in precision is observed. Notably, substantial improvements are manifested in the CRF and BERT models. In contrast, the RuBERT and RuCYBERT models exhibit negligible enhancement, thereby alluding to the potential significance of fine-tuning prior to the application of data augmentation techniques. This observation propels further inquiry into the intricate interplay between fine-tuning and data augmentation, and their collective impact on model performance.

In Longpre et al.~\cite{Longpre2020HowEI} the authors assessed the efficacy of backtranslation alongside a suite of easy data augmentation (EDA) techniques across various NLP tasks, employing six diverse datasets and three distinct transformer architectures: BERT, RoBERTa~\cite{Liu2019RoBERTaAR}, and XLNet~\cite{10.5555/3454287.3454804}. Despite not directly addressing Named Entity Recognition, the research holds significance. Their findings reveal an inconsistent enhancement in quality by the examined text augmentation techniques on transformer models. This inconsistency underscores the critical role of the scale of pre-training in models constructed using transformers when contemplating the application of data augmentation. Furthermore, Longpre et al. propose a potential preference for task-related augmentation over task-agnostic methods, exemplified by the use of back-translation for machine translation tasks.

There has been limited advancement in low-resource Named Entity Recognition (NER) settings. This work aims to address this gap by conducting further experiments and presenting additional results. Utilizing established NER architectures and augmentation techniques, such as those outlined in  Section~\ref{sec:background}, the experiments are designed in accordance with methods used in current research. In particular, there is a notable lack of research concerning the impact of the number of augmented sentences on NER models. Although one study has touched upon this issue~\cite{Liu2020LongtailDE}, it is of a smaller scale compared to the present work. The amount of augmentation is an important \emph{hyperparameter} that must be set before training the model, as it can introduce noise and affect the quality of the dataset. By investiganting the effect of different numbers of augmented sentences, as described in Section~\ref{sec:exp-setup}, this work aims to provide insights into how this parameter influences NER model quality.

\section{Experimental Setup}\label{sec:exp-setup}

In this section, we present the details of the experiments we carried out on the effectiveness of data augmentation in different low-resource domain datasets.

\subsection{Datasets}

In our experiments, we used four different datasets. Two are from the biomedical domain, BioCreative V CDR Task~\cite{10.1093/database/baw068} and i2b2-2010~\cite{10.1136/amiajnl-2011-000203}, one is from the materials science domain, MaSciP~\cite{mysore-etal-2019-materials}, and the last one is from the legal domain, JusBrasil\footnote{Jusbrasil \url{https://sobre.jusbrasil.com.br/} is the largest online platform in Brazil for legal search.}. Other articles on NER models~\cite{dai-adel-2020-analysis, ijcai2022p590} previously used i2b2-2010 and MaSciP. For all datasets, if a train-dev-test split was not originally provided, we used 15\% of the original data for testing and 85\% for training, and we used 15\% of the training data as validation data. Table~\ref{tab:datasets} summarizes the number of sentences, tokens, and entities in each dataset.

\begin{table*}[t]
    \caption{Datasets used in the experiment}
    \scalebox{0.92}{
    \begin{tabular}{lcccccc}
        \toprule
        \multirow{2}{*}{Dataset} & \multicolumn{3}{c}{Training Set} & \multicolumn{3}{c}{Validation Set} \\
         & Sentences & Tokens & Entities & Sentences & Tokens & Entities \\
        \midrule
        \textbf{i2b2-2010} & 13867 & 127151 & 3 & 2448 & 22390 & 3 \\
        \textbf{MaSciP} & 2253 & 53718 & 21 & 138 & 3548 & 20 \\
        \textbf{BioCreative V CDR} & 1000 & 116913 & 2 & 1000 & 115965 & 2 \\
        \textbf{JusBrasil} & 1817 & 123597 & 1 & 321 & 21279 & 1 \\
        \bottomrule
    \end{tabular}}
 \label{tab:datasets}
\end{table*}

\begin{table*}[t]
    \scalebox{0.92}{
    \begin{tabular}{lcccc}
        \toprule
        \multirow{2}{*}{Dataset} & \multicolumn{3}{c}{Testing Set} & \multirow{2}{*}{Language} \\
         & Sentences & Tokens & Entities &  \\
        \midrule
        \textbf{i2b2-2010} & 27625 & 267249 & 3 & English \\
        \textbf{MaSciP} & 182 & 4066 & 21 & English \\
        \textbf{BioCreative V CDR} & 1000 & 122838 & 2 & English \\
        \textbf{JusBrasil} & 378 & 25521 & 1 & Portuguese (BR) \\
        \bottomrule
    \end{tabular}}
 \label{tab:datasets}
\end{table*}

To assess the impact of data augmentation in various low-resource scenarios, we selected different sentence subsets from each training dataset: 50, 150, and 500 sentences, aligning with prior research~\cite{dai-adel-2020-analysis, ijcai2022p590}. Additionally, subsets amounting to 25\%, 50\%, and 75\% of the original dataset size are chosen, provided these subsets exceed 500 sentences. The validation and testing datasets remain unaltered throughout the evaluation.

\subsection{Data Augmentation}

We used two techniques in our experiments, \emph{Mention Replacement (MR)} and \emph{Contextual Word Replacement (CWR)}, both described in Section~\ref{sec:background}. We applied two small changes to MR. First, we replaced the mentions in the sentences with new mentions only seen in the training subsets. This does not apply to the full dataset. This was done because we did not want the technique to use additional information from external sources, which would be the discarded part of the dataset for each subset. Second, we selected a random number of tokens of the new mention to produce more variability in the dataset. For augmented sentences with CWR, we applied the technique only to words whose tags are not of the target entity types of the dataset, i.e. the annotated tag is ``O''. In this way, we assume that BERT is not highly specialized in the low-resource language domains of the datasets, replacing only the words that require less technical expertise. By applying these two augmentation techniques separately and with the aforementioned changes, we can generate augmented sentences by applying data augmentation only to the tokens of target entity types (MR), and sentences where augmentation was applied only to tokens of tag "O" (CWR).

\subsection{Setup}

We selected two popular architectures used to build NER models, Bi-LSTM+CRF~\cite{DBLP:journals/corr/HuangXY15} and BERT~\cite{devlin-etal-2019-bert}. We used SciBERT~\cite{beltagy-etal-2019-scibert}, a fine-tuned version of BERT for scientific texts, for datasets from the biomedical and materials science domains, and regular multilingual BERT for the other datasets. For each dataset and its subsets, we gradually increased the amount of augmented data in the training, ranging from 0\% (original data without augmented data) to 500\% (original data plus 5 times the amount of original data as augmented data). Finally, we used validation loss as a parameter to interrupt model training if it stopped decreasing for 5 epochs. This setup yields 784 groups of models, considering the combination of all datasets and subsets, architectures, and augmentation techniques and amounts. Each group consists of 10 models trained from scratch, which we evaluated in the testing data, in order to check model variance. We calculated the average F1-score of the models of each group, which we present in Section~\ref{sec:results}.

\newcommand{\footexp}{\footnote{\url{https://github.com/arttorres0/augmented-ner-model}}}

All datasets and codes used in the experiments are publicly available to encourage the carrying out of more experiments\footexp.

\section{Results and Discussion}\label{sec:results}

\balance

We now present the results of experiments performed with the datasets described in Section~\ref{sec:exp-setup} to study the effects of different amounts of data augmentation on the quality of NER models.

\subsection{General Results}
\label{sec:general-results}

\begin{figure*}
    \centering
    \begin{tabular}{cc}
       \includegraphics[width=180pt]{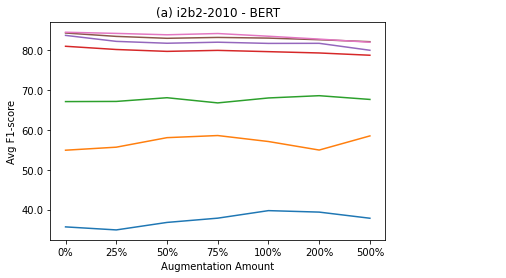} & \includegraphics[width=180pt]{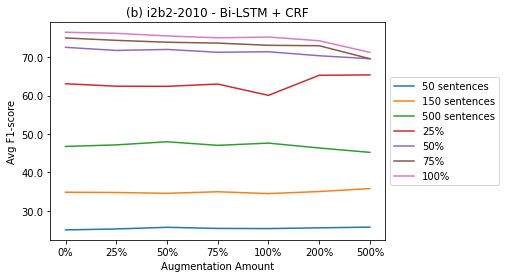} \\
       \includegraphics[width=180pt]{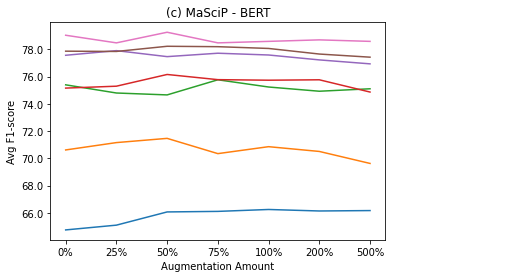} & \includegraphics[width=180pt]{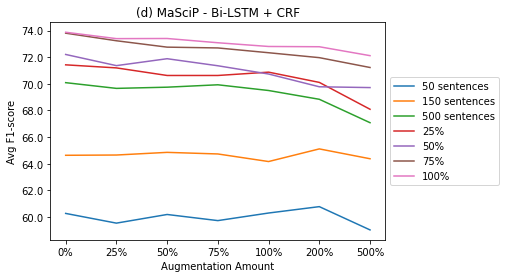} \\
       \includegraphics[width=180pt]{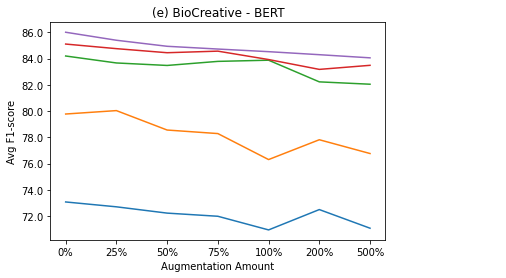} & \includegraphics[width=180pt]{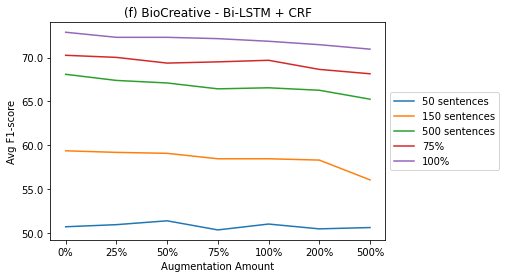} \\
       \includegraphics[width=180pt]{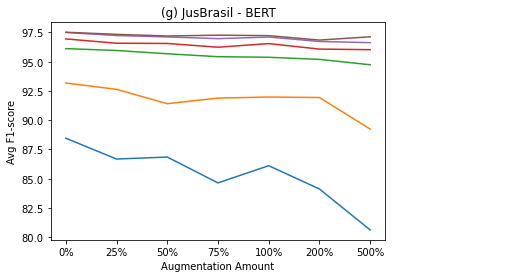} & \includegraphics[width=180pt]{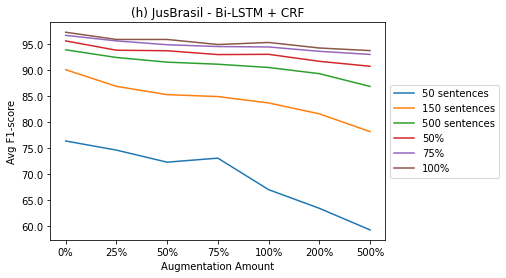} \\
    \end{tabular}
    \caption{Effects of different amounts of augmented examples (MR) on average F1-score by model and dataset.}
    \label{fig:results-graphs-mr}
\end{figure*}

\begin{figure*}
    \centering
    \begin{tabular}{cc}
       \includegraphics[width=180pt]{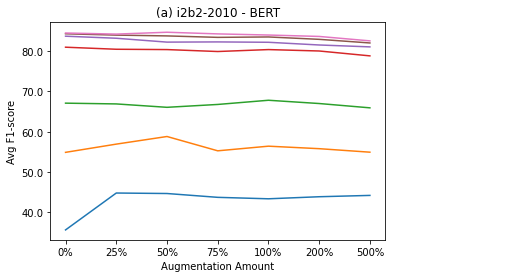} & \includegraphics[width=180pt]{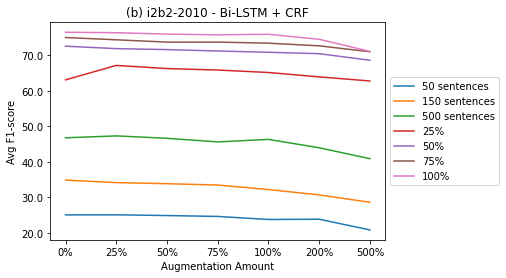} \\
       \includegraphics[width=180pt]{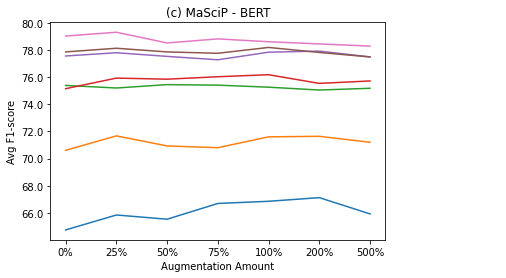} & \includegraphics[width=180pt]{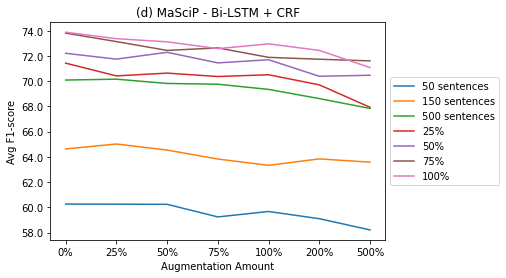} \\
       \includegraphics[width=180pt]{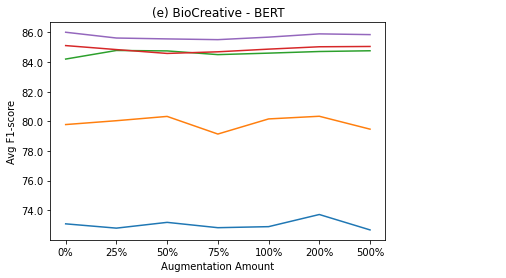} & \includegraphics[width=180pt]{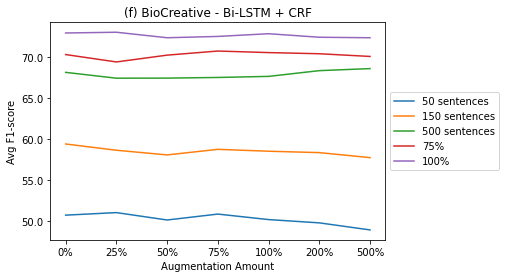} \\
       \includegraphics[width=180pt]{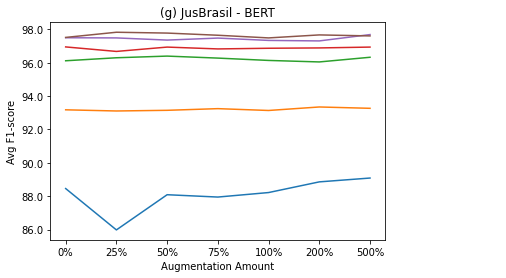} & \includegraphics[width=180pt]{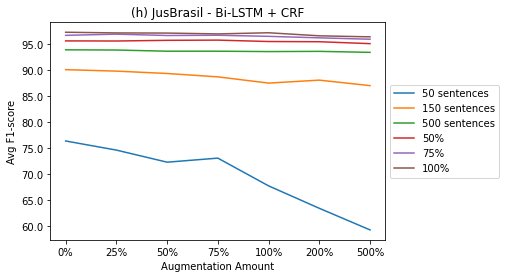} \\
    \end{tabular}
    \caption{Effects of different amounts of augmented examples (CWR) on average F1-score by model and dataset.}
    \label{fig:results-graphs-cwr}
\end{figure*}

In Figures~\ref{fig:results-graphs-mr} and ~\ref{fig:results-graphs-cwr}, we plot a distinct graph for each dataset and architecture pair, using, respectively, MR and CWR as the augmentation technique. For each of these pairs, we plot the average F1-score of each 10-model group as a function of the amount of data augmentation used. Each curve in the graphs corresponds to a subset of the full dataset, both in terms of percentage (25\%, 50\%, 75\%, and 100\%) and in terms of the absolute number of sentences (50, 150, and 500). As detailed in Section~\ref{sec:exp-setup}, depending on the size of the dataset, not all percentages and absolute numbers of sentences were considered.

Looking at Figures~\ref{fig:results-graphs-mr} and ~\ref{fig:results-graphs-cwr}, we notice that the BERT models outperform the Bi-LSTM+CRF models for almost all datasets. This is consistent with previous results in the literature that compare BERT and Bi-LSTM+CRF models (e.g., \cite{dai-adel-2020-analysis, ijcai2022p590}). The smaller-sized subsets of \emph{JusBrasil} dataset are an exception. This is likely because the scale of pre-training for multilingual BERT favors other languages more than Brazilian Portuguese.

We observe a modest increase in model quality for lower-sized subsets before the model starts to reduce the F1-score value, when analyzing models augmented with MR. However, for the full dataset and other larger subsets, the average F1-score has not increased, maintaining or losing quality. The only exceptions to this are \emph{MaSciP} (BERT), in which every subset benefits from data augmentation, and \emph{JusBrasil} (Bi-LSTM+CRF), in which every subset loses model quality for every amount of data augmentation. This reduction in F1-score may be a consequence of the introduction of invalid augmented data, which had their original ground-truth labels altered by the augmentation technique. This suggests that data augmentation should be used with caution, as the injection of high noise into the training dataset may cause models to overfit these invalid instances. This trade-off between diversity and validity has been reported in previous research~\cite{dai-adel-2020-analysis, 10.5555/3495724.3496249}. These inconsistencies could also be due to other factors, such as the characteristics of the datasets or the parameters of the augmentation techniques. We believe that identifying these factors is an important area for future research.

Models augmented with CWR show similar results. However, we notice that more models benefit from data augmentation. This is an indicator that this technique produces more data variability than MR in the studied datasets, which makes sense, as BERT can suggest several different tokens. We also have more possible target replacements ("O" tokens) than when applying MR, which targets non-"O" tokens. On the other hand, we observe a more inconsistent pattern of improvement in F1-score across different subset sizes. This is also a consequence of the higher variability introduced by BERT, but on the negative side, where there is more noise introduction than actual valid augmented examples in the datasets.

\subsection{Detailed Results}

Tables~\ref{tab:results-table-i2b2} -~\ref{tab:results-table-jusbrasil-cwr} present a different perspective of the above results. We show the average F1-score for each subset and each model without data augmentation, and indicate which amount of augmentation yields the best models in average. This summarization yields 100 averages. We also determine whether the augmentation technique produces statistically significant differences between the model without augmentation and the best augmented model, determined by paired T-student tests. This is only if the best amount of augmentation is greater than 0\%. In these tables, we indicate the training size and, for each model, we present the F1-score of the model with no augmentation (column "F1 No Aug"), the F1-score of the best model (column "F1 Best"), the absolute difference in F1-score between these two (column "$\Delta$"), and which augmentation amount yields the best model (column "Aug Amount"). Column "T-student Test" indicates the result of a T-student test, if "$\Delta$" is nonzero, where "S" indicates that the difference is statistically significant, and "NS" means that it is not.

The pattern observed in the results of Section~\ref{sec:general-results} can also be observed in these tables, where we see some smaller subsets benefiting from data augmentation, while larger subsets never improve. While we refer to ‘smaller’ datasets throughout this paper, it’s important to note that the term is relative and depends on the specific context and domain. For instance, in our experiments, we found that data augmentation was particularly beneficial for subsets of the datasets with fewer than 500 sentences, such as the 50-sentence subsets of the BioCreative or JusBrasil datasets. However, this threshold may vary for other datasets or domains, and further research is needed to determine the optimal size for applying data augmentation techniques. Another observation is that there is no optimal predetermined amount of data augmentation, probably due to the random introduction of noise by the augmentation techniques. Because of this, we can conclude that NER practitioners should test different amounts of augmented examples in their models. 

Table~\ref{tab:results-summary-table} summarizes the content of the previous tables, presenting all models that achieved statistically significant improvements by data augmentation. We sorted these models by improvement, in descending order. We note that 10 of the 100 best models, or 10\%, were found to present statistically significant improvements in the average F1-score. The conclusions above are also evident here, that is, first, smaller subsets improve with data augmentation, unlike larger subsets, and, second, in spite of these promising results, there is not an augmentation amount that yields the best results.

We also observe that, for the datasets studied in this experiment, CWR yields more improved models than MR, that is, 7 versus 4. This can be an indication that CWR should be prioritized when selecting data augmentation techniques for NER models. We can also see that two of the four datasets benefit from data augmentation, where i2b2-2010 and MaSciP have 5 improved models each. Although i2b2-2010 and BioCreative are similar datasets in terms of domain and number of entities, models of the first dataset are improved by data augmentation, unlike models of the second dataset, which never improved. Finally, BERT models are improved more than Bi-LSTM+CRF models, that is, 9 versus 2. This can be an indication that data augmentation suits BERT-based NER models better than Bi-LSTM+CRF-based ones.

The results obtained in these experiments show that there is an unfilled potential for the improvement of low-resource NER models. With the exception of the JusBrasil dataset, where the average F1-score for the models that use full datasets is around 98\%, all models of the other datasets have below 90\% F1-score. Compared to other datasets of high-resource domains studied in the literature, where the quality of the model is higher~\cite{yadav-bethard-2018-survey,li-etal-2020-dice,automated-concatenation,external-context-retriev,zhong-chen-2021-frustratingly}, there is still a margin for improvement of low-resource NER models.



\begin{table*}[h]
    \caption{0\% and best augmentation (MR) average F1-scores and applied paired T-student tests for i2b2 dataset}
    \scalebox{0.73}{
    \begin{tabular}{c|ccccc|ccccc}
        \toprule
        \textbf{} & \multicolumn{5}{c}{\textbf{i2b2-2010 (BERT)}} & \multicolumn{5}{|c}{\textbf{i2b2-2010 (Bi-LSTM+CRF)}} \\
        Training  & F1     & F1    & $\Delta$ & Aug   & T-student  & F1   & F1   & $\Delta$ & Aug  & T-student \\
        Size      & No Aug & Best  &          & Amount & Test      & No Aug & Best &          & Amount        & Test \\
        \midrule
        \textbf{50 sent.} & 35.65 & 39.73 & 11.44\% & 100\% & NS & 25.01 & 25.74 & 2.92\% & 500\% & S \\
        \textbf{150 sent.} & 54.89 & 58.57 & 6.70\% & 75\% & S & 34.83 & 35.77 & 2.70\% & 500\% & NS \\
        \textbf{500 sent.} & 67.09 & 68.57 & 2.21\% & 200\% & NS & 46.75 & 47.97 & 2.61\% & 50\% & NS \\
        \textbf{25\%} & 80.97 & 80.97 & 0 & 0\% & - & 63.08 & 65.39 & 3.66\% & 500\% & NS \\
        \textbf{50\%} & 83.7 & 83.7 & 0 & 0\% & - & 72.57 & 72.57 & 0 & 0\% & - \\
        \textbf{75\%} & 84.3 & 84.3 & 0 & 0\% & - & 75.01 & 75.01 & 0 & 0\% & - \\
        \textbf{Full} & 84.49 & 84.49 & 0 & 0\% & - & 76.5 & 76.5 & 0 & 0\% & - \\
        \bottomrule
    \end{tabular}
    \label{tab:results-table-i2b2}}
\end{table*}

\clearpage

\begin{table*}[!]
    \caption{0\% and best augmentation (MR) average F1-scores and applied paired T-student tests for MaSciP dataset}
    \scalebox{0.73}{
    \begin{tabular}{c|ccccc|ccccc}
        \toprule
        \textbf{} & \multicolumn{5}{c}{\textbf{MaSciP (BERT)}} & \multicolumn{5}{|c}{\textbf{MaSciP (Bi-LSTM+CRF)}} \\
        Training  & F1     & F1    & $\Delta$ & Aug   & T-student  & F1   & F1   & $\Delta$ & Aug  & T-student \\
        Size      & No Aug & Best  &          & Amount & Test      & No Aug & Best &          & Amount        & Test \\
        \midrule
        \textbf{50 sent.} & 64.74 & 66.24 & 2.32\% & 100\% & S & 60.26 & 60.77 & 0.85\% & 200\% & NS \\
        \textbf{150 sent.} & 70.61 & 71.46 & 1.20\% & 50\% & NS & 64.63 & 65.11 & 0.74\% & 200\% & NS \\
        \textbf{500 sent.} & 75.39 & 75.76 & 0.49\% & 75\% & NS & 70.09 & 70.09 & 0 & 0\% & - \\
        \textbf{25\%} & 75.15 & 76.15 & 1.33\% & 50\% & S & 71.43 & 71.43 & 0 & 0\% & - \\
        \textbf{50\%} & 77.56 & 77.9 & 0.44\% & 25\% & NS & 72.21 & 72.21 & 0 & 0\% & - \\
        \textbf{75\%} & 77.86 & 78.22 & 0.46\% & 50\% & NS & 73.81 & 73.81 & 0 & 0\% & - \\
        \textbf{Full} & 79.03 & 79.25 & 0.28\% & 50\% & NS & 73.88 & 73.88 & 0 & 0\% & - \\
        \bottomrule
    \end{tabular}
    \label{tab:results-table-mascip}}
\end{table*}

\begin{table*}[!]
    \caption{0\% and best augmentation (MR) average F1-scores and applied paired T-student tests for BioCreative dataset}
    \scalebox{0.73}{
    \begin{tabular}{c|ccccc|ccccc}
        \toprule
        \textbf{} & \multicolumn{5}{c}{\textbf{BioCreative (BERT)}} & \multicolumn{5}{|c}{\textbf{BioCreative (Bi-LSTM+CRF)}} \\
        Training  & F1     & F1    & $\Delta$ & Aug   & T-student  & F1   & F1   & $\Delta$ & Aug  & T-student \\
        Size      & No Aug & Best  &          & Amount & Test      & No Aug & Best &          & Amount        & Test \\
        \midrule
        \textbf{50 sent.} & 73.08 & 73.08 & 0 & 0\% & - & 50.7 & 51.38 & 1.34\% & 50\% & NS \\
        \textbf{150 sent.} & 79.78 & 80.04 & 0.33\% & 25\% & NS & 59.37 & 59.37 & 0 & 0\% & - \\
        \textbf{500 sent.} & 84.2 & 84.2 & 0 & 0\% & - & 68.1 & 68.1 & 0 & 0\% & - \\
        \textbf{75\%} & 85.11 & 85.11 & 0 & 0\% & - & 70.27 & 70.27 & 0 & 0\% & - \\
        \textbf{Full} & 86.01 & 86.01 & 0 & 0\% & - & 72.9 & 72.9 & 0 & 0\% & - \\
        \bottomrule
    \end{tabular}
    \label{tab:results-table-biocreative}}
\end{table*}

\begin{table*}[!]
    \caption{0\% and best augmentation (MR) average F1-scores and applied paired T-student tests for JusBrasil dataset}
    \scalebox{0.73}{
    \begin{tabular}{c|ccccc|ccccc}
        \toprule
        \textbf{} & \multicolumn{5}{c}{\textbf{JusBrasil (BERT)}} & \multicolumn{5}{|c}{\textbf{JusBrasil (Bi-LSTM+CRF)}} \\
        Training  & F1     & F1    & $\Delta$ & Aug   & T-student  & F1   & F1   & $\Delta$ & Aug  & T-student \\
        Size      & No Aug & Best  &          & Amount & Test      & No Aug & Best &          & Amount        & Test \\
        \midrule
        \textbf{50 sent.} & 88.46 & 88.46 & 0 & 0\% & - & 76.34 & 76.34 & 0 & 0\% & - \\
        \textbf{150 sent.} & 93.18 & 93.18 & 0 & 0\% & - & 90.1 & 90.1 & 0 & 0\% & - \\
        \textbf{500 sent.} & 96.12 & 96.12 & 0 & 0\% & - & 93.93 & 93.93 & 0 & 0\% & - \\
        \textbf{50\%} & 96.95 & 96.95 & 0 & 0\% & - & 95.65 & 95.65 & 0 & 0\% & - \\
        \textbf{75\%} & 97.5 & 97.5 & 0 & 0\% & - & 96.72 & 96.72 & 0 & 0\% & - \\
        \textbf{Full} & 97.52 & 97.52 & 0 & 0\% & - & 97.32 & 97.32 & 0 & 0\% & - \\
        \bottomrule
    \end{tabular}
    \label{tab:results-table-jusbrasil}}
\end{table*}


\begin{table*}[!]
    \caption{0\% and best augmentation (CWR) average F1-scores and applied paired T-student tests for i2b2 dataset}
    \scalebox{0.73}{
    \begin{tabular}{c|ccccc|ccccc}
        \toprule
        \textbf{} & \multicolumn{5}{c}{\textbf{i2b2-2010 (BERT)}} & \multicolumn{5}{|c}{\textbf{i2b2-2010 (Bi-LSTM+CRF)}} \\
        Training  & F1     & F1    & $\Delta$ & Aug   & T-student  & F1   & F1   & $\Delta$ & Aug  & T-student \\
        Size      & No Aug & Best  &          & Amount & Test      & No Aug & Best &          & Amount        & Test \\
        \midrule
        \textbf{50 sent.} & 35.65 & 44.8 & 25.67\% & 25\% & S & 25.01 & 25.04 & 0.12\% & 25\% & NS \\
        \textbf{150 sent.} & 54.89 & 58.82 & 7.16\% & 50\% & S & 34.83 & 34.83 & 0 & 0\% & - \\
        \textbf{500 sent.} & 67.09 & 67.81 & 1.07\% & 100\% & NS & 46.75 & 47.28 & 1.13\% & 25\% & NS \\
        \textbf{25\%} & 80.97 & 80.97 & 0 & 0\% & - & 63.08 & 67.15 & 6.45\% & 25\% & S \\
        \textbf{50\%} & 83.7 & 83.7 & 0 & 0\% & - & 72.57 & 72.57 & 0 & 0\% & - \\
        \textbf{75\%} & 84.3 & 84.3 & 0 & 0\% & - & 75.01 & 75.01 & 0 & 0\% & - \\
        \textbf{Full} & 84.49 & 84.68 & 0.22\% & 50\% & NS & 76.5 & 76.5 & 0 & 0\% & - \\
        \bottomrule
    \end{tabular}
    \label{tab:results-table-i2b2-cwr}}
\end{table*}

\begin{table*}[!]
    \caption{0\% and best augmentation (CWR) average F1-scores and applied paired T-student tests for MaSciP dataset}
    \scalebox{0.73}{
    \begin{tabular}{c|ccccc|ccccc}
        \toprule
        \textbf{} & \multicolumn{5}{c}{\textbf{MaSciP (BERT)}} & \multicolumn{5}{|c}{\textbf{MaSciP (Bi-LSTM+CRF)}} \\
        Training  & F1     & F1    & $\Delta$ & Aug   & T-student  & F1   & F1   & $\Delta$ & Aug  & T-student \\
        Size      & No Aug & Best  &          & Amount & Test      & No Aug & Best &          & Amount        & Test \\
        \midrule
        \textbf{50 sent.} & 64.74 & 67.12 & 3.68\% & 200\% & S & 60.26 & 60.26 & 0 & 0\% & - \\
        \textbf{150 sent.} & 70.61 & 71.67 & 1.50\% & 25\% & S & 64.63 & 65.02 & 0.60\% & 25\% & NS \\
        \textbf{500 sent.} & 75.39 & 75.45 & 0.08\% & 50\% & NS & 70.09 & 70.16 & 0.10\% & 25\% & NS \\
        \textbf{25\%} & 75.15 & 76.18 & 1.37\% & 100\% & S & 71.43 & 71.43 & 0 & 0\% & - \\
        \textbf{50\%} & 77.56 & 77.92 & 0.46\% & 200\% & NS & 72.21 & 72.29 & 0.11\% & 50\% & NS \\
        \textbf{75\%} & 77.86 & 78.19 & 0.42\% & 100\% & NS & 73.81 & 73.81 & 0 & 0\% & - \\
        \textbf{Full} & 79.03 & 79.31 & 0.35\% & 25\% & NS & 73.88 & 73.88 & 0 & 0\% & - \\
        \bottomrule
    \end{tabular}
    \label{tab:results-table-mascip-cwr}}
\end{table*}

\begin{table*}[!]
    \caption{0\% and best augmentation (CWR) average F1-scores and applied paired T-student tests for BioCreative dataset}
    \scalebox{0.73}{
    \begin{tabular}{c|ccccc|ccccc}
        \toprule
        \textbf{} & \multicolumn{5}{c}{\textbf{BioCreative (BERT)}} & \multicolumn{5}{|c}{\textbf{BioCreative (Bi-LSTM+CRF)}} \\
        Training  & F1     & F1    & $\Delta$ & Aug   & T-student  & F1   & F1   & $\Delta$ & Aug  & T-student \\
        Size      & No Aug & Best  &          & Amount & Test      & No Aug & Best &          & Amount        & Test \\
        \midrule
        \textbf{50 sent.} & 73.08 & 73.71 & 0.86\% & 200\% & NS & 50.7 & 51.01 & 0.61\% & 25\% & NS \\
        \textbf{150 sent.} & 79.78 & 80.34 & 0.70\% & 200\% & NS & 59.37 & 59.37 & 0 & 0\% & - \\
        \textbf{500 sent.} & 84.2 & 84.78 & 0.69\% & 25\% & NS & 68.1 & 68.56 & 0.68\% & 500\% & NS \\
        \textbf{75\%} & 85.11 & 85.11 & 0 & 0\% & - & 70.27 & 70.7 & 0.61\% & 75\% & NS \\
        \textbf{Full} & 86.01 & 86.01 & 0 & 0\% & - & 72.9 & 72.99 & 0.12\% & 25\% & NS \\
        \bottomrule
    \end{tabular}
    \label{tab:results-table-biocreative-cwr}}
\end{table*}

\begin{table*}[!]
    \caption{0\% and best augmentation (CWR) average F1-scores and applied paired T-student tests for JusBrasil dataset}
    \scalebox{0.73}{
    \begin{tabular}{c|ccccc|ccccc}
        \toprule
        \textbf{} & \multicolumn{5}{c}{\textbf{JusBrasil (BERT)}} & \multicolumn{5}{|c}{\textbf{JusBrasil (Bi-LSTM+CRF)}} \\
        Training  & F1     & F1    & $\Delta$ & Aug   & T-student  & F1   & F1   & $\Delta$ & Aug  & T-student \\
        Size      & No Aug & Best  &          & Amount & Test      & No Aug & Best &          & Amount        & Test \\
        \midrule
        \textbf{50 sent.} & 88.46 & 89.09 & 0.71\% & 500\% & NS & 76.34 & 76.34 & 0 & 0\% & - \\
        \textbf{150 sent.} & 93.18 & 93.35 & 0.18\% & 200\% & NS & 90.1 & 90.1 & 0 & 0\% & - \\
        \textbf{500 sent.} & 96.12 & 96.4 & 0.29\% & 50\% & NS & 93.93 & 93.93 & 0 & 0\% & - \\
        \textbf{50\%} & 96.95 & 96.95 & 0 & 0\% & - & 95.65 & 95.8 & 0.16\% & 75\% & NS \\
        \textbf{75\%} & 97.5 & 97.69 & 0.19\% & 500\% & NS & 96.72 & 96.95 & 0.24\% & 25\% & NS \\
        \textbf{Full} & 97.52 & 97.83 & 0.32\% & 25\% & NS & 97.32 & 97.32 & 0 & 0\% & - \\
        \bottomrule
    \end{tabular}
    \label{tab:results-table-jusbrasil-cwr}}
\end{table*}


\begin{table*}[t]
    \caption{Summary of models improved by augmentation}
    \scalebox{0.8}{
    \begin{tabular}{lllrrl}
        \toprule
        \multicolumn{1}{c}{\textbf{Dataset}} & \multicolumn{1}{c}{\textbf{Training Size}} & \multicolumn{1}{c}{\textbf{Model}} & \multicolumn{1}{c}{\textbf{Improvement}} & \multicolumn{1}{c}{\textbf{Aug amount}} & \multicolumn{1}{c}{\textbf{Aug Technique}} \\
        \midrule
        i2b2-2010 & 50 sentences & BERT & 25.67\% & 25\% & CWR \\
        i2b2-2010 & 150 sentences & BERT & 7.16\% & 50\% & CWR \\
        i2b2-2010 & 150 sentences & BERT & 6.70\% & 75\% & MR \\
        i2b2-2010 & 25\% & Bi-LSTM + CRF & 6.45\% & 25\% & CWR \\
        MaSciP & 50 sentences & BERT & 3.68\% & 200\% & CWR \\
        i2b2-2010 & 50 sentences & Bi-LSTM + CRF & 2.92\% & 500\% & MR \\
        MaSciP & 50 sentences & BERT & 2.32\% & 100\% & MR \\
        MaSciP & 150 sentences & BERT & 1.50\% & 25\% & CWR \\
        MaSciP & 25\% & BERT & 1.37\% & 100\% & CWR \\
        MaSciP & 25\% & BERT & 1.33\% & 50\% & MR \\
        \bottomrule
        \end{tabular}
    \label{tab:results-summary-table}}
\end{table*}



\section{Conclusion and Future Work}\label{sec:conclusion}

In this paper, we investigate the effectiveness of two representative data augmentation techniques, \emph{Mention Replacement}~\cite{dai-adel-2020-analysis} and \emph{Contextual Word Replacement}~\cite{jiao-etal-2020-tinybert}, when used for NER tasks on low-resource domain datasets, which contain long-tail entities. In our study, we use two popular neural architectures for NER, Bi-LSTM+CRF and BERT, to generate several different NER models with distinct configurations. For training, these configurations used subsets of varying sizes of the original datasets.    

Specifically, we generated 784 groups of 10 models, from which we measure the average F1-score and pick the best one for each subset. This yields 100 averages, 10 of which are considered statistically significant improvements after using text augmentation. As these improved models were those trained using smaller subsets, our study further supports the notion that data augmentation for NER models is more effective for smaller training sets than for larger ones. We also verify that there is no predetermined optimal number of augmented instances that will certainly improve the model. Not only that, augmentation also reduces model quality in several other cases. Thus, NER practitioners should test different amounts, especially if the augmentation technique is prone to introduce spurious instances in the training data. We also show that, for the datasets studied in this paper, CWR yields models that are better than MR, and that data augmentation suits BERT NER models better than Bi-LSTM+CRF.

Our work on augmentation for NER in low-resource domains indicates that this landscape can still be improved in both the case of top-notch but high-demanding architectures such as BERT, or in the case of competitive but less-demanding architectures such as Bi-LSTM+CRF. While our study provides valuable insights into the effectiveness of data augmentation techniques for Named Entity Recognition (NER) on low-resource domains, it also highlights several limitations that warrant further investigation.

Firstly, our findings show that data augmentation can introduce noise into the training data, which can potentially degrade the quality of the NER models. This is particularly evident in our experiments with larger subsets of the datasets, where data augmentation often resulted in lower average F1-scores compared to models trained without augmented data. This suggests that while data augmentation can help to increase the quantity of training data, it may also compromise its quality, leading to overfitting or poor generalization to unseen data.

To mitigate this issue, future research could explore more sophisticated or controlled augmentation techniques that minimize the introduction of noise. For example, one could consider techniques that take into account the semantic consistency between the original and augmented sentences, or that adjust the degree of augmentation based on the complexity or diversity of the dataset.

Secondly, our study assumes that all entities are equally important and should be augmented at the same rate. However, in many real-world applications, some entities may be more critical or rare than others, and therefore might benefit more from augmentation. Future work could investigate differential augmentation strategies that prioritize certain entity types based on their importance or rarity.

Lastly, our study focuses on two specific NER architectures (Bi-LSTM+CRF and BERT) and two text augmentation techniques (MR and CWR). While these methods are widely used and have shown promising results in many NLP tasks, they may not be optimal or applicable for all types of NER problems or domains. Future research should consider exploring other architectures and techniques, as well as hybrid or ensemble methods that combine the strengths of multiple approaches.

\bmhead{Acknowledgments}
 This work was partially supported by Diffbot Inc., Jusbrasil Fellowship Program, Coordenacao de Aperfeicoamento de Pessoal de Nivel Superior (Grant No. 001), Amazonas State Research Support Foundation - FAPEAM - through the POSGRAD project, and Conselho Nacional de Desenvolvimento Cientifico e Tecnologico (Grants No. 307248/2019-4 and 310573/2023-8).

\section*{Declarations}
\bmhead{Funding} Project funding gently provided by Diffbot, JusBrasil, CAPES, FAPEAM and CNPq.
\bmhead{Conflicts of Interest}
The authors have no conflicts of interest to declare that are relevant to the content of this article.
\bmhead{Data Availability}
All datasets and subsets used in this project and all notebooks generated during the experiments are available in the following repository: https://github.com/arttorres0/augmented-ner-model.

\bibliography{sn-bibliography}

\end{document}